\documentclass{article}

\usepackage[dblblindworkshop, final]{neurips_2026}
\workshoptitle{Workshop on Resource-Aware Agentic AI}
\usepackage[utf8]{inputenc} % allow utf-8 input
\usepackage[T1]{fontenc}    % use 8-bit T1 fonts
\usepackage{hyperref}       % hyperlinks
\usepackage{url}            % simple URL typesetting
\usepackage{booktabs}       % professional-quality tables
\usepackage{amsmath}        % cases, \text, align
\usepackage{amsfonts}       % blackboard math symbols
\usepackage{amssymb}        % \checkmark (feature-survivor table)
\usepackage{nicefrac}       % compact symbols for 1/2, etc.
\usepackage{microtype}      % microtypography
\usepackage{xcolor}         % colors
\usepackage{float}          % [H] hard placement for the results tables
\newcommand{\sysname}{Sentinel}  % system name: change once here
\usepackage{graphicx}       % \includegraphics for figures

\title{Trajectory-Derived Confidence for Reliable, Resource-Aware Clinical Text-to-SQL Agents}

\author{%
  Mincheol Daniel Song \quad Joshua Ward \quad Jake Jung \quad Guang Cheng \\
  University of California, Los Angeles \\
  \texttt{danielsong05@g.ucla.edu, joshuaward@ucla.edu,} \\
  \texttt{jakejung27@g.ucla.edu, guangcheng@stat.ucla.edu} \\
}

\newcommand{\pans}{\hat{p}_{\mathrm{ans}}}
\newcommand{\pcorr}{\hat{p}_{\mathrm{corr}}}
\newcommand{\rs}[1]{\mathrm{RS}(#1)}

\begin{document}
\maketitle
\begin{abstract}
LLM agents for clinical text-to-SQL applications reason autonomously over multiple steps but cannot assess whether their own reasoning or outputs can be trusted. In high leverage applications such as healthcare, this presents a critical risk where system mistakes can be costly. These reliability failures are also resource failures: an incorrect reasoning trajectory spends computation budget on outputs that must be discarded. We introduce \sysname{}, a trajectory-derived, classifier-based confidence layer that analyzes an agent's reasoning, code and database outputs to decide at three points whether to stop: refusing unanswerable questions before the agent runs, halting doomed trajectories mid-run, and withholding untrustworthy answers at delivery. Here, utilizing Chow’s rule, we optimize decisions under the EHRSQL shared task’s Reliability Score, which penalizes incorrect answers given a utility weighting, and find on the benchmark EHRSQL that \sysname{} raises this score from $+0.08$ to $+0.24$ when mistakes have a low utility weighting, well above the $+0.03$ earned by refusing every question, with delivered-answer accuracy rising from 54\% to 69\% as coverage falls from 84\% to 55\%. At higher stakes the agents we test rarely answer reliably enough to deliver, and \sysname{} detects this on its own, abstaining to that same $+0.03$ where the unmonitored agent scores $-3.02$. The same stopping decisions cut computation: at low stakes, where the system still answers, \sysname{} eliminates 13--28\% of agent steps at little or no reliability cost. 

\end{abstract}

\section{Introduction}
\label{sec:intro}

\begin{figure}[t]
  \centering
  \includegraphics[width=0.72\linewidth]{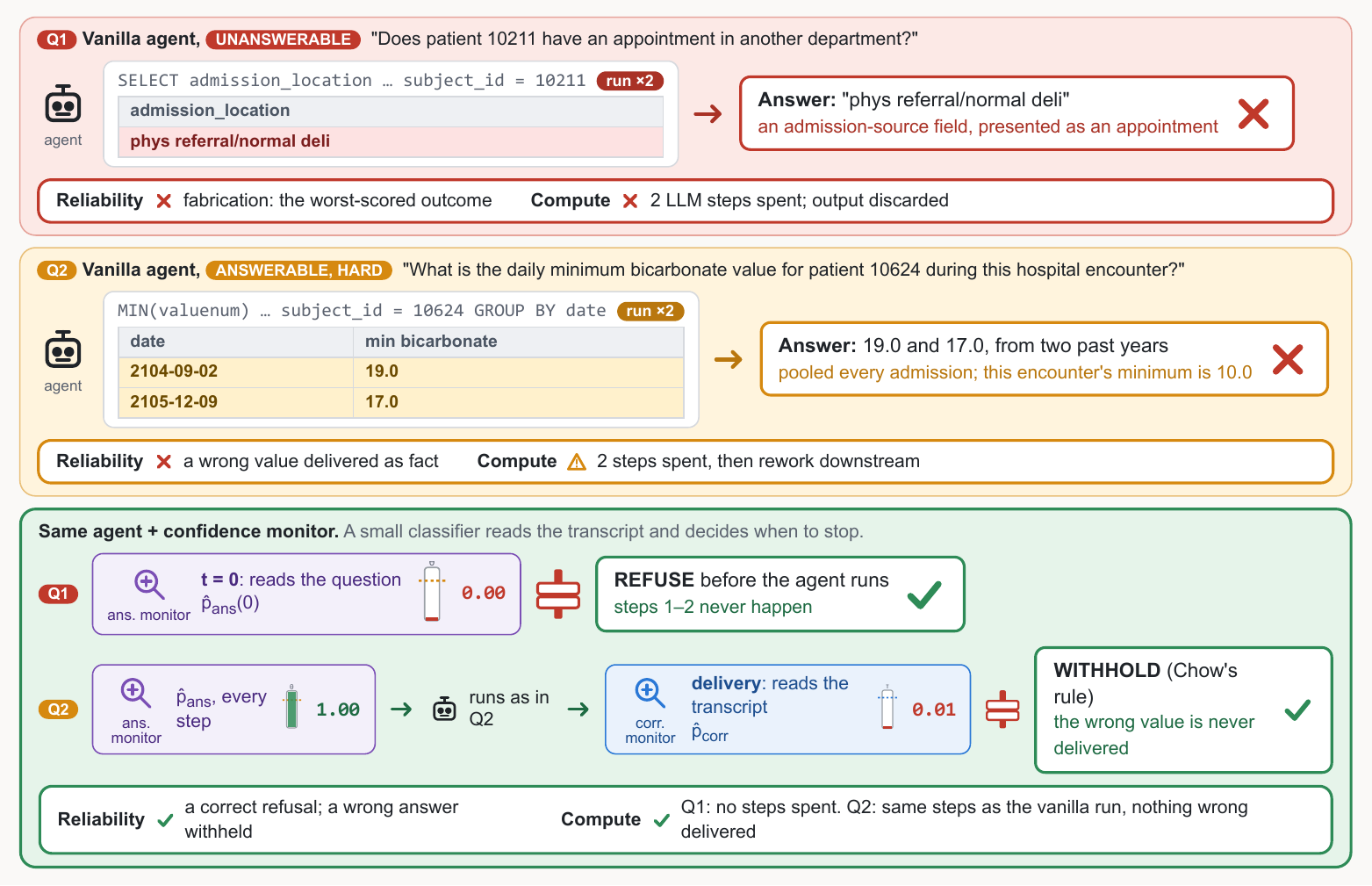}
  \caption{\textbf{Diagram of failure modes} Q1: on an
  unanswerable question the vanilla ReAct agent queries an unrelated
  admission field twice and commits it as an answer. Q2: on a question
  scoped to the current encounter, it pools every past admission and
  delivers a wrong value as fact. Bottom panel: the answerability head
  (purple) refuses Q1 before the agent runs; the correctness head
  (blue) withholds Q2's wrong answer at delivery; each panel's
  Compute row prices the steps spent or spared. Gauges show actual
  out-of-fold monitor estimates against thresholds $\theta$ and
  $\tau$. }
  \label{fig:motivation}
  \vspace{-1.1em}
\end{figure}

Recent advances in LLMs have shifted tabular reasoning from its traditional Table QA framework
\citep{pasupat2015compositional, zhong2017seq2sql, yu2018spider} to increasingly autonomous agents that can reason through their decision-making. Rather than producing a final answer in one pass, these agents can issue SQL queries or table operations, inspect the intermediate results, and revise their next steps based on that feedback \citep{yao2023react,
wang2024chainoftable, kwok2026tabqaworld}. With
this
autonomy comes a new risk: an agent given full freedom to reason must
also discern for itself whether that reasoning and resulting response can be trusted.

In clinical settings, the cost of a wrong answer is high, explicit, and often greater than the cost of declining \citep{swaminathan2024selective}. There are two failure modes we consider that carry distinct harms. The first is \emph{incorrectness}, where the database has enough information to answer the question, but the agent returns an answer that differs from the correct result. For example, the agent may return a wrong lab value, medication count, or cohort size, but is delivered in the same fluent, confident register as a correct one. Clinicians are rarely positioned to audit the SQL that produced it, and tend to accept confident machine output even when it is wrong \citep{goddard2012automation}. The second failure mode is \emph{unanswerability}, where the database lacks the information needed to answer the question. In this case, the correct behavior is to abstain rather than return an unsupported answer. Yet agents routinely fabricate: in our runs, even the strongest agent we test (\texttt{GPT-4o};
Section~\ref{sec:experiments}) commits an answer to a third of
unanswerable questions.

These reliability failures are simultaneously \emph{resource}
failures. Each agent step is a full, costly LLM call \citep{chen2023frugalgpt} over a growing context,
so a trajectory that should never have started wastes computation and a user's time, producing an output the system
must discard. Responses to unanswerable questions show some of the highest costs because the correct
action, abstention, is available at step zero; everything afterward
is waste by construction. In our experimentation, a vanilla GPT-4o
agent on average spends 2.5 reasoning steps per unanswerable question before
failing. When it fabricates an answer, 
it takes 3.35 steps on average, 58\% more than the
2.12 steps it takes on average when it correctly refuses
(per-agent step economics in Appendix~\ref{app:peragent}). Knowing \emph{when to stop}, before the first step,
mid-trajectory, or at delivery, is therefore one decision problem
serving two goals at once: not delivering costly mistakes, and not
paying for them.

The obstacle is that the agent cannot reliably assess its own correctness. LLMs can be poorly calibrated when verbalizing confidence, and may express high confidence even when their answers are wrong \citep{xiong2023uncertainty}. More fundamentally, the model's stated reasoning may not reflect the actual reasons behind its final answer, making chain-of-thought an unreliable indicator of model behavior \citep{turpin2023unfaithful,lanham2023faithfulness}. The interaction trajectory, however, includes externally grounded signals that the agent does not control, including executed SQL, error messages, empty result sets, and failed
schema references. A separate monitor can learn how these externally observed patterns correlate with eventual success across past trajectories, allowing it to estimate reliability without depending on the agent’s own explanation or confidence estimate.

Figure~\ref{fig:motivation} demonstrates both failures modes and
\sysname{}, a \emph{trajectory-derived confidence layer} wrapping an
unmodified ReAct agent \citep{yao2023react}. Two small text classifiers read the
agent's own transcript: an \emph{answerability} head powers an exit gate that halts 
trajectories on unanswerable questions, while a
\emph{correctness} head provides a calibrated estimate that stops doomed trajectories 
mid-run and, via Chow's rule \citep{chow1970}, withholds
untrustworthy answers at delivery. Rather than training the agent to
be more correct, the system delivers less costly mistakes by
refraining from answering, and stops paying for trajectories that
will not end in a delivered correct answer.

Our contributions can be summarized as follows:
\begin{itemize}\setlength\itemsep{0.15em}
  \item \textbf{Stakes-aware reliability.}
    We formulate reliability for clinical text-to-SQL agents as a
    stakes-dependent decision problem, where the cost of an incorrect answer
    can exceed the cost of abstaining.

    \item \textbf{Trajectory-derived monitoring.}
    \sysname{} introduces a confidence layer that scores the agent's evolving trajectory rather than relying solely on its final output or self-reported confidence (Figure~\ref{fig:pipeline}). The monitor operates before execution, during reasoning, and at delivery, allowing it to refuse unanswerable questions, stop low-confidence trajectories, and withhold unreliable final answers across five agent models.

    \item \textbf{Resource-aware early stopping.}
    We show that the same confidence estimates can reduce computation by
    stopping trajectories that are unlikely to produce a useful answer,
    reducing unnecessary reasoning steps and latency while preserving
    reliability at several operating points.
\end{itemize}

\section{Related work}
\label{sec:related}

\noindent\textbf{Text-to-SQL over EHRs and abstention.}
Early text-to-SQL systems focused on mapping natural-language 
questions to executable SQL, including Seq2SQL
\citep{zhong2017seq2sql}
and the Spider benchmark for complex cross-domain querying
\citep{yu2018spider}. Later work introduced multi-step reasoning, decomposition, and execution feedback to improve performance on complex queries
\citep{pourreza2023dinsql,wang2018execution}. In the clinical domain, EHRSQL
\citep{lee2022ehrsql} introduced unanswerable questions, making
abstention a first-class capability for clinical text-to-SQL. TrustSQL
\citep{lee2024trustsql} introduced the penalty-based Reliability
Score, later the official metric of the EHRSQL 2024 shared task
\citep{lee2024ehrsqlshared}, and evaluated pipelines that split
abstention into unanswerable-question detection and SQL-error
detection, the same decomposition our two heads adopt.
Shared-task systems such as ProbGate \citep{kim2024probgate} filter
unanswerable questions by thresholding token log-probabilities; our
training-free telemetry baseline uses the same token-level signals.
The nearest reliability baseline is the IDK filter of \citet{broecker2026idk}, 
a trainable uncertainty filter for function-calling agents whose IDK 
score coincides with $\rs{1}$ on answerable
questions; Appendix~\ref{app:peragent} compares gains on that shared scale.
Unlike these approaches, \sysname{} derives confidence from the agent's reasoning 
trajectory and applies that signal at multiple decision
points to refuse unanswerable questions, stop low-confidence trajectories
early, and withhold unreliable final answers.

\smallskip

\noindent\textbf{Selective prediction and agent confidence.}
Classification with a reject option
dates to \citet{chow1957optimum, chow1970}. Modern selective-prediction methods allow a model to abstain on low-confidence inputs and study the resulting trade-off between coverage, the fraction of inputs on which the model makes a prediction, and selective risk, the error on those retained predictions \citep{geifman2017selective, kamath2020selective, hendrickx2024survey}.
We apply this framing at two distinct checkpoints
of an agentic trajectory rather than to a single prediction. First, we use confidence during execution to see whether an ongoing trajectory should be stopped early when unlikely to succeed. Second, we use confidence at delivery time to decide whether the completed answer should be returned or withheld.
Uncertainty signals have been drawn from token probabilities and
entropy, verbalized confidence, and learned confidence heads
\citep{kadavath2022know, xiong2023uncertainty}. Our trajectory
features draw from three different approaches: token-level confidence estimation \citep{kadavath2022know, kim2024probgate, somov2025confidence}, execution-guided text-to-SQL \citep{wang2018execution}, and clause-level SQL verification \citep{gong2025sqlens} or tool-grounded verification of table-reasoning steps \citep{zou2025tattoo}. We adapt signals from these three methods into features computed over the agent's trajectory, with per-feature sources provided in Appendix~\ref{app:features}.
Closest to us is work on process supervision for intermediate table-reasoning
states, where learned or deterministic reward signals score each step of a
trajectory and are used to guide, rerank, or halt the reasoning policy, or as a training reward \citep{zou2025tattoo, kwok2026tabalign, kwok2026tabrouge}. In contrast, we never steer or select the agent's reasoning: we only observe its trajectory, predict whether it will end in a correct delivered answer, and use that prediction to stop or withhold under a reliability objective.
\section{Method}
\label{sec:method}

% ------------------------------------------------------------------
\subsection{Problem formulation}
\label{sec:formulation}
\begin{figure}[t]
  \centering
  \includegraphics[width=0.78\linewidth]{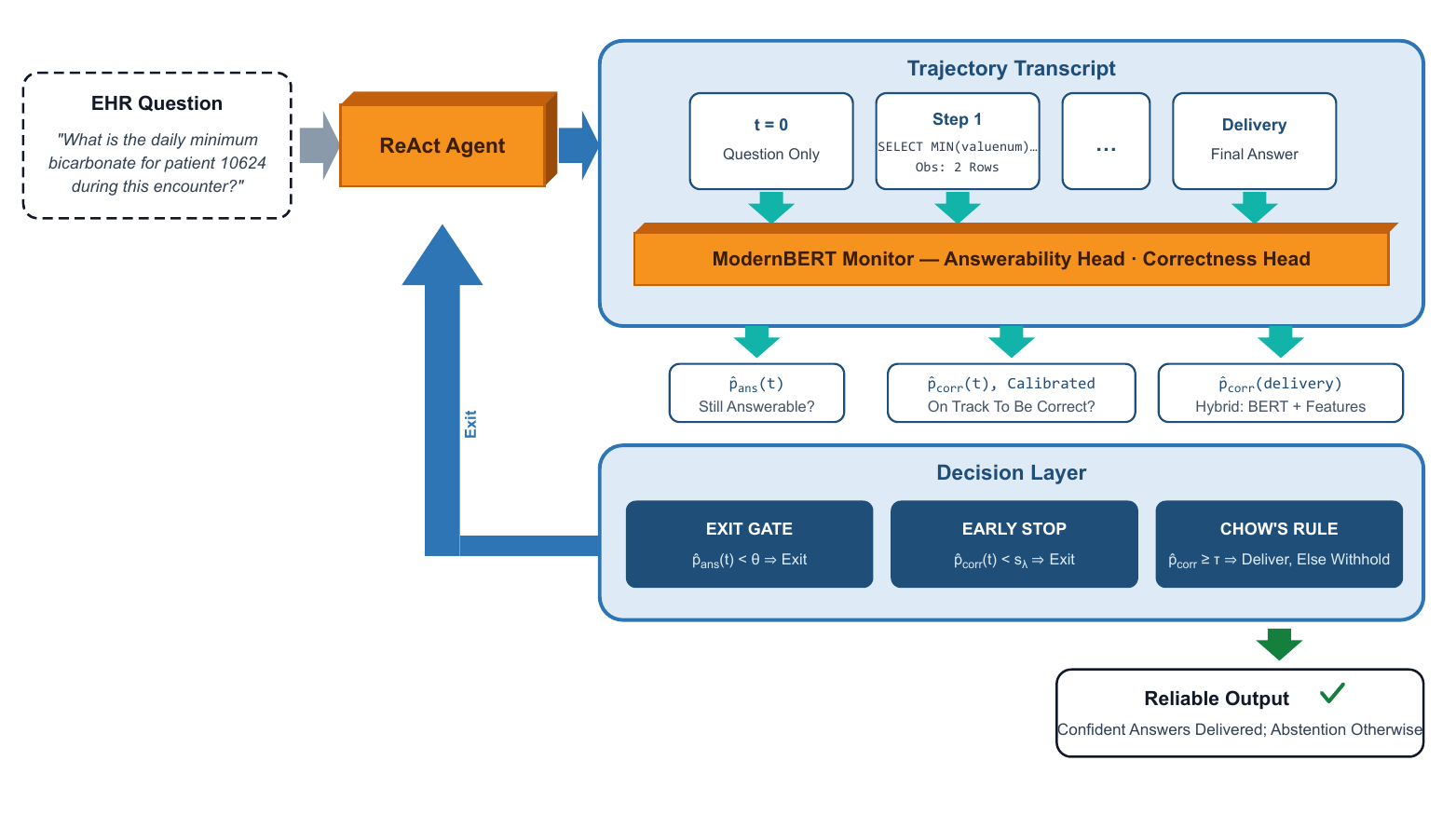}
  \caption{\textbf{The \sysname{} pipeline.} A frozen ReAct agent
  appends each reasoning/SQL/result step to a serialized transcript.
  Two ModernBERT heads score every prefix, including $t{=}0$, so an
  unanswerable question can be refused before the agent runs a single
  step. Answerability drives the
exit gate ($\theta$); calibrated correctness drives early stopping
($s_\lambda$) and Chow's rule at delivery ($\tau$).}
  \label{fig:pipeline}
  \vspace{-1.1em}
\end{figure}

\paragraph{Setting.} A question $q$ is posed against a database $D$. Let $a \in \{0,1\}$ indicate whether $q$ is \emph{answerable} from $D$; $a$ is not observed by the system. Formally, $q$ is answerable iff there exists a SQL query over $D$ whose results yield the correct answer; otherwise the question is deemed unanswerable (e.g., a coffee shop's order database cannot answer \emph{which paper introduced attention mechanisms} under any query). Thus $a$ is determined by $(q, D)$ alone, rather than by intuition. An agent processes $q$ as a trajectory of states $s_0, s_1, \dots, s_T$: the initial state $s_0 = (q)$ contains only the question, and each subsequent state appends one reasoning step, $s_t = s_{t-1} \oplus (r_t, u_t, o_t)$, where $r_t$ is the agent's reasoning text, $u_t$ an (optional) SQL query, and $o_t$ the observed execution result. Let $\hat{y}$ denote the agent's committed output at the terminal state, outputting either an answer or the abstention symbol $\bot$ (the agent's \texttt{FINAL: null}). When the agent answers and $a = 1$, let $y \in \{0,1\}$ indicate whether $\hat{y}$ is correct. Two events harm the user: answering an unanswerable question, $\{a = 0,\ \hat{y} \neq \bot\}$, and delivering a wrong answer to an answerable question, $\{a = 1,\ \hat{y} \neq \bot,\ y = 0\}$; the first event wastes all compute, since $\bot$ was available at $s_0$.

\paragraph{Reliability Score (RS).} Both failure events carry real cost in high-stakes environments. In a clinical setting, a delivered mistake can directly harm the user, and in every setting it wastes the compute spent producing it. Our method adopts the Reliability Score (RS) of TrustSQL~\citep{lee2024trustsql}, the official metric of the EHRSQL 2024 shared task~\citep{lee2024ehrsqlshared}, which rewards correct answers and correct refusals and penalizes delivered mistakes at a stakes parameter $\lambda$:
\begin{equation}
\rs{\lambda} \;=\;
\begin{cases}
+1 & \hat{y} \neq \bot,\ a = 1,\ y = 1
     \quad \text{(correct answer)}\\
+1 & \hat{y} = \bot,\ a = 0
     \quad \text{(correct refusal)}\\
0 & \hat{y} = \bot,\ a = 1
     \quad \text{(missed answer)}\\
-\lambda & \text{otherwise}
     \quad \text{(wrong or unanswerable answered)}
\end{cases}
\label{eq:rs}
\end{equation}
The stakes are evaluated at $\lambda \in \{1, 4, 9\}$ so that the corresponding optimal delivery thresholds (Section~\ref{sec:decisions}) cover low, medium, and high stakes: at $\lambda = 1$, a delivered mistake cancels a point earned from a
correct answer or refusal, while at $\lambda = 9$, a single mistake accounts for nine accumulated points.

\subsection{Decisions}
\label{sec:decisions}
Our goal is to reduce costly mistakes under $\rs{\lambda}$ while spending as little compute as possible. To accomplish this, we wrap the vanilla agent in a confidence layer (Figure~\ref{fig:pipeline}) that may intervene on the
agent's own trajectory in two different ways. At every state $s_t$ ($t = 0, \dots, T$), the layer may \textsc{exit} the trajectory and output $\bot$ at any given moment $t$, disregarding the remaining $T - t$ steps. At delivery, if the agent committed to an answer, the layer may \textsc{withhold}, outputting $\bot$ instead of $\hat{y}$. Both interventions convert a potential $-\lambda$ output into one worth at least $0$ (exactly $0$ if the question in fact was answerable, $+1$ for correctly abstaining), since the act of abstention is never penalized under $\rs{\lambda}$.

\paragraph{Optimal delivery: Chow's rule.} For the intervention \textsc{withhold}, the optimal rule requires no tuning; it follows directly from the $\rs{\lambda}$ payoffs. Suppose the agent has
committed an answer, $\hat{y} \neq \bot$, at the terminal state. Let $p = P(a = 1,\ y = 1 \mid s_T, \hat{y})$ denote the probability that the committed answer scores $+1$ (note that $1 - p$ covers both
failure modes: a wrong answer, or an unanswerable question answered).
Delivering has expected utility $p(+1) + (1-p)(-\lambda)$, while
withholding is worth $0$ on an answerable question, so delivering is optimal iff
\begin{equation}
p \;\ge\; \frac{\lambda}{1+\lambda} \;=:\; \tau_\lambda,
\label{eq:chow}
\end{equation}
which represents Chow's rule \citep{chow1970}. Plugging in the stakes
$\lambda \in \{1, 4, 9\}$, we get delivery thresholds
$\tau_\lambda = 0.5,\ 0.8,\ 0.9$. A chance $\varepsilon$ that the question is unanswerable raises
the exact bar to $(\lambda + \varepsilon)/(1+\lambda)$; the gate keeps
$\varepsilon$ small, and the exact bar changes no reported score by
more than $0.01$ (Appendix~\ref{app:abstain}). Chow's rule is only optimal for
honest probabilities, so the estimate of $p$ is calibrated before
thresholding (Section~\ref{sec:inference}). The intervention
\textsc{exit} has no closed form. Stopping a trajectory that
Chow's rule would later withhold leaves the score unchanged and saves
the unrun steps, but an intermediate estimate can also stop a
trajectory that would have been delivered, so early stoppage can move
the score in either direction. Because $\rs{\lambda}$ assigns no value
to the steps saved, the \textsc{exit} triggers are set empirically
(Section~\ref{sec:inference}).
% ------------------------------------------------------------------
\subsection{Estimating confidence from trajectories}
\label{sec:heads}

The decisions of Section~\ref{sec:decisions} require two conditional
probabilities, each available at every intermediate state: the
probability that the question is answerable,
$\pans(t) = P(a = 1 \mid s_t)$, and the probability that the
trajectory will end in a delivered correct answer,
$\pcorr(t) = P(\hat{y} \neq \bot,\ a = 1,\ y = 1 \mid s_t)$. Note
that answerability is part of the event that $\pcorr$ predicts; by
chain rule, $\pcorr(t)$ factors into $\pans(t)$ times correctness
given answerability, so a trajectory on a likely unanswerable
question scores low even when its SQL looks plausible. Either
probability can justify an \textsc{exit}, since both instances lead
to wasted compute. \textsc{Withhold} uses the terminal estimate
$\pcorr(T) = P(a = 1,\ y = 1 \mid s_T, \hat{y})$, which is exactly
the quantity $p$ referenced in Section~\ref{sec:decisions}.

\paragraph{Two trajectory-reading heads.} We estimate both
probabilities with text classifiers
(ModernBERT-base~\citep{warner2024modernbert};
Appendix~\ref{app:encoder}) that read a plain-text rendering of the
state $s_t$: the question $q$, followed by each step's reasoning,
query, and result $(r_i, u_i, o_i)$ for $i \le t$. The rendering is
left-truncated so that the most recent steps are always retained
(2{,}048 tokens in training, 4{,}096 at scoring; fewer than 0.1\% of
transcripts reach either cap). The answerability head estimates
$\pans(t)$ and is trained with unanswerable ($a = 0$) as the positive
class. Its score at $t = 0$ is computed from $s_0 = (q)$ alone,
before the agent runs a single step, and serves as a pre-filter. The
correctness head is trained on answered answerable questions
($\hat{y} \neq \bot$, $a = 1$) and therefore estimates correctness
conditional on answerability; Section~\ref{sec:inference} describes
how its output is calibrated to the joint quantity $\pcorr(t)$ that
the decisions require. Neither head consumes hand-engineered
features; hand-crafted trajectory features enter only the
delivery-time estimator selection of Section~\ref{sec:inference}.

\paragraph{Prefix training.} Each collected trajectory is expanded
into prefix snapshots at $t = 0, 1, \dots, T$, and every snapshot is
labeled with the trajectory's eventual outcome. The final answer
line is included only in the terminal snapshot, so each prefix
contains exactly what was knowable at that state. All training-pool
estimates reported in this paper are out-of-fold: 5-fold
cross-validation grouped by question template, so all prefixes of a
trajectory, and all questions sharing a template, stay within one
fold. 
% ------------------------------------------------------------------
\subsection{Inference procedure}
\label{sec:inference}

At inference, the confidence layer applies, in order:

\textbf{(1) \textsc{exit} gate.} \textsc{Exit} triggers the exact moment $\pans(t)$ drops below a gate threshold $\theta$, and the t = 0 reading can refuse a question before the agent runs a single step. $\theta$ is predetermined as the largest threshold that wrongly exits at most 0.5\% of answerable questions on the calibration pool. The gate is the system's defense against unanswerable questions.

\textbf{(2) Early \textsc{exit} (resource-aware mode).} Mid run, \textsc{exit} also triggers when $\pcorr(t)$,
calibrated by isotonic regression (a monotone, nonparametric recalibration map;~\citealp{zadrozny2002transforming}) fit on the
calibration pool's prefix snapshots, falls below a stop threshold $s_\lambda$. Since $\rs{\lambda}$ assigns no value to steps (Section~\ref{sec:decisions}), $s_\lambda$ is chosen empirically on the calibration pool to maximize $\rs{\lambda}$, ties broken toward more steps saved.

\textbf{(3) Chow's rule at delivery.} Once the trajectory reaches the final step $T$, only committed answers are checked. The correctness estimate is calibrated by a separate isotonic regression fit on every answer delivered in the calibration pool, with answers to unanswerable questions counted as incorrect, so the calibrated value estimates the joint probability $p$ of Section~\ref{sec:decisions}. Deliver iff
the calibrated estimate
$\hat{p}_{\mathrm{corr}} \ge \tau_\lambda = \lambda/(1+\lambda)$
(Eq.~\eqref{eq:chow}) and otherwise \textsc{withhold}. The exact
delivery estimator is selected on the calibration pool (Appendix~\ref{app:features}); we report calibration error before and after.

All thresholds, calibration maps, and the estimator selection are
frozen on the calibration pool, and held-out pools are evaluated exactly
once.

\section{Experiments}
\label{sec:experiments}
\subsection{Setup}
\label{sec:setup}

We compare the vanilla agent against the full confidence layer at
stakes $\lambda \in \{1, 4, 9\}$, and evaluate its parts separately:
the two checks in isolation (gate-only and Chow-only), and the
resource-aware mode (early \textsc{exit}), which trades delivered
answers for saved compute.

\paragraph{Benchmark.} We use EHRSQL \citep{lee2022ehrsql} on
MIMIC-III \citep{johnson2016mimic}, a benchmark of natural-language
questions over a de-identified ICU database that includes questions
unanswerable from the schema. From the official train and validation
splits we construct three pools with no question template shared
between pools: a train pool (6{,}179 questions; 231 unanswerable), a
calibration pool (2{,}160; 71), and a held-out pool (2{,}101; 60),
evaluated exactly once. The official test split is left untouched.

\paragraph{Agents.} The vanilla agent is a ReAct loop
\citep{yao2023react} in which the model, at each step, produces a
thought and optionally SQL, observes the execution result, and
finally commits an answer or abstains. Five agent models run under
an identical harness and protocol, namely \texttt{GPT-4o}
\citep{openai2024gpt4o} and \texttt{GPT-4o-mini}
\citep{openai2024gpt4omini},
\texttt{Qwen2.5-Coder-7B} \citep{hui2024qwen25coder},
\texttt{Llama-3.1-8B} \citep{grattafiori2024llama3}, and
\texttt{Phi-4-mini} \citep{microsoft2025phi4mini}. Full experimental
details, including the agent prompt and decoding configuration, are
in Appendices~\ref{app:prompts} and~\ref{app:repro}.

\paragraph{Correctness labels.} EHRSQL's official metric is exact
execution match, which is strict for an agent that writes its own
final query: the executed result can express the gold answer in a
different form, such as \texttt{90} against a gold \texttt{90.0}. We
therefore use a semantic label, in which an LLM judge
(\texttt{GPT-4o-mini}, itself one of the evaluated agents) deems an
answer correct if its executed result is equivalent to the gold
query's executed result; exact matches are auto-correct. The
semantic label is primary throughout; labeling details and
strict-label scoring are in Appendices~\ref{app:labels}
and~\ref{app:abstain}.

\paragraph{Delivery estimator.} A pre-registered rule selects, per
agent, between the correctness head alone and a gradient-boosted
hybrid of the head with ten trajectory features, whichever scores
higher on the calibration pool, preferring the simpler within
$0.005$. Feature definitions are in Appendix~\ref{app:features}.

\subsection{Reliability}
\label{sec:results}

Table~\ref{tab:models} gives the cross-agent result. The vanilla agents become increasingly unreliable as the cost of an incorrect answer rises. Even the strongest,
\texttt{GPT-4o}, is wrong on 46\% of the answers it delivers, so its
reliability score barely clears zero at $\lambda{=}1$ and falls to
$-3.02$ at $\lambda{=}9$, and the weakest agents reach $-5.72$. The
reference point is a system that refuses every question, which
scores $+0.03$ at every stakes level because abstention is never
penalized. At $\lambda{=}1$ the full system clears it for every
agent, significantly for four of the five: \texttt{GPT-4o} delivers
55\% of questions at 69\% accuracy and gains $+0.21$. At
$\lambda{=}4$ and $\lambda{=}9$ the system delivers little or nothing
and matches always abstaining within noise. We read this as a gating
result: at these stakes no agent produces enough reliable answers to
deploy, and the layer reaches that verdict from the trajectories
alone, where the vanilla agent scores $-3.02$.

\begin{figure}[t]
  \centering
  \includegraphics[width=\linewidth]{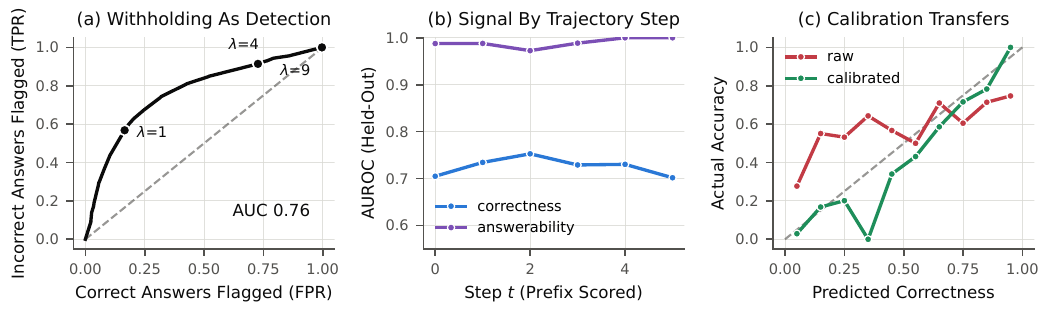}
  \caption{\textbf{Behavior of the confidence layer.} (a)~The full
system as a single withhold classifier, with dots marking Chow's
rule at $\tau_\lambda = 0.5, 0.8, 0.9$. (b)~AUROC of each head
scored on the prefix at step $t$; later points cover only the longer
runs still going.
(c)~Reliability diagram of the
correctness head before and after the isotonic map fit on the
calibration pool.}
  \label{fig:results}
  \vspace{-1.1em}
\end{figure}

Of the two checks, Chow's rule at delivery accounts for nearly all
of the gain. The gate on its own moves $\rs{\lambda}$ by $+0.01$ to
$+0.14$, since unanswerable questions are only 3\% of the pool; at
the 20--33\% rates of EHRSQL's own evaluation splits the gate's share
grows and the $\lambda{=}1$ gain over always abstaining remains
(Appendix~\ref{app:abstain}).
Viewed as a single binary detector of answers that should be
withheld, the layer separates wrong deliveries from right ones at
$0.76$ AUC on held-out committed answers
(Figure~\ref{fig:results}a).

\begin{table}
  \caption{\textbf{Cross-agent reliability.} Vanilla acc.\ = accuracy of
  the answers the raw agent delivers on the held-out pool (answers to
  unanswerable questions count as wrong); monitor quality =
  correctness / answerability head AUROC on the same pool.
  \emph{Always abstain} refuses every question and scores the
  unanswerable fraction at any $\lambda$. $\dagger$: gain over always
  abstaining whose 95\% bootstrap interval excludes zero
  (Appendix~\ref{app:abstain}). Per-agent early-\textsc{exit} results
  are in Table~\ref{tab:es-agents}.}
  \label{tab:models}
  \centering
  \footnotesize
  \setlength{\tabcolsep}{4pt}
  \begin{tabular}{l c cc ccc ccc}
    \toprule
    & Vanilla & \multicolumn{2}{c}{Monitor AUROC} & \multicolumn{3}{c}{Vanilla $\rs{\lambda}$} & \multicolumn{3}{c}{Full System $\rs{\lambda}$} \\
    \cmidrule(lr){3-4} \cmidrule(lr){5-7} \cmidrule(lr){8-10}
    Agent & Acc. & Corr. & Ans. & $1$ & $4$ & $9$ & $1$ & $4$ & $9$ \\
    \midrule
    \texttt{GPT-4o-mini} & $48\%$ & $0.787$ & $0.923$ & $-0.02$ & $-1.34$ & $-3.53$ & $+0.21^{\dagger}$ & $+0.08$ & $+0.04$ \\
    \texttt{GPT-4o}      & $54\%$ & $0.762$ & $0.981$ & $+0.08$ & $-1.08$ & $-3.02$ & $+0.24^{\dagger}$ & $+0.02$ & $+0.03$ \\
    \texttt{Llama-3.1-8B} & $31\%$ & $0.707$ & $0.903$ & $-0.23$ & $-1.61$ & $-3.91$ & $+0.06$ & $+0.03$ & $+0.03$ \\
    \texttt{Qwen2.5-Coder-7B} & $22\%$ & $0.920$ & $0.959$ & $-0.46$ & $-2.43$ & $-5.72$ & $+0.10^{\dagger}$ & $+0.03$ & $+0.03$ \\
    \texttt{Phi-4-mini}  & $8.7\%$ & $0.921$ & $0.987$ & $-0.46$ & $-2.03$ & $-4.64$ & $+0.04^{\dagger}$ & $+0.03$ & $+0.03$ \\
    \midrule
    \emph{Always abstain} & --- & --- & --- & \multicolumn{3}{c}{---} & $+0.03$ & $+0.03$ & $+0.03$ \\
    \bottomrule
  \end{tabular}
  \vspace{-1.1em}
\end{table}

\subsection{Resource savings}
\label{sec:resource}

Early \textsc{exit} applies the delivery logic sequentially,
stopping a trajectory as soon as its calibrated per-step correctness
estimate falls below the stop threshold. Table~\ref{tab:resource}
shows what this buys on \texttt{GPT-4o}. At $\lambda{=}1$, where
the system still answers over half of the questions, it eliminates
13\% of all agent steps with no loss in reliability, and 13--28\%
across agents at a cost of at most $0.01$ $\rs{1}$
(Appendix~\ref{app:peragent}). At $\lambda{=}4$ and $\lambda{=}9$ the saving rises to
42\% (up to 70\% on other agents), but there the system delivers
almost nothing, so this is compute recovered from trajectories whose
answers would have been withheld; refusing every question outright
would recover all of it. At a measured 1.9 seconds per agent step, each
avoided step is also one fewer LLM call and one fewer live SQL query
against the clinical database.

\begin{table}[H]
  \caption{\textbf{Resource savings} (\texttt{GPT-4o}). Agent steps,
  generated tokens, and wall-clock time never spent. The monitor's two
  ModernBERT-base passes per step are not subtracted. Token
  accounting in Appendix~\ref{app:tokens}.}
  \label{tab:resource}
  \centering
  \small
  \begin{tabular}{l ccc}
    \toprule
    System & Steps Saved & Gen.\ Tokens Saved & Total Time Saved \\
    & ($\lambda{=}1/4/9$) & ($\lambda{=}1/4/9$) & ($\lambda{=}1/4/9$) \\
    \midrule
    + \textsc{exit} gate / + Chow delivery & $1\%$ & $1\%$ & $3$\,min \\
    + early \textsc{exit} & $\mathbf{13\%\,/\,42\%\,/\,42\%}$ & $16\%\,/\,47\%\,/\,47\%$ & $29$\,min\,/\,$1.5$\,h\,/\,$1.5$\,h \\
    \bottomrule
  \end{tabular}
  \vspace{-1.1em}
\end{table}
\section{Discussion}
\label{sec:discussion}

The signal is in the trajectory, not the question. Reading the
question alone, the correctness head scores $0.705$ AUROC for
\texttt{GPT-4o}; at delivery it reaches $0.762$, and on
\texttt{Qwen2.5-Coder-7B} a single step of execution evidence lifts
it from $0.742$ to $0.898$ (Figure~\ref{fig:results}b; Appendix~\ref{app:abstain}). Training-free telemetry does
not suffice; mean token entropy reaches only $0.565$ for
correctness on \texttt{GPT-4o-mini}, mirroring \citet{somov2025confidence}, who find
selective classifiers catch unanswerable-question errors more
readily than wrong-but-plausible SQL. The raw head is also
overconfident, and the isotonic map fit on the calibration pool
holds on the held-out pool (Figure~\ref{fig:results}c; Appendix~\ref{app:bins}), as does the
deployed estimate (held-out ECE $0.03$--$0.07$ vs.\ $0.25$ raw), so
the thresholds operate on honest probabilities. For \texttt{GPT-4o} the calibrated delivery
estimate exceeds $0.9$ on only three held-out questions, which is why the system abstains almost entirely at
$\lambda{=}9$. The layer also adapts to how each agent fails.
\texttt{Phi-4-mini} fabricates answers to half of its unanswerable
questions, and the layer's verdict is near-total refusal, the
appropriate response. \texttt{Qwen2.5-Coder-7B} answers 84\% of
questions at 22\% accuracy, and the layer still delivers 21\% of
questions at 68\% accuracy at $\lambda{=}1$.

How monitorable an agent is tracks not its capability but how
legibly it fails. The two loudest failers, whose trajectories fill
with execution errors and fast fabrications, are the easiest to read
($0.920$, $0.921$), while \texttt{GPT-4o}, whose rare failures
resemble its successes, and \texttt{Llama-3.1-8B}, which meanders
through the longest trajectories of the five, are the hardest
($0.762$, $0.707$). As agents improve and their failures get
quieter, the signal this approach reads may fade.

\paragraph{Limitations.}
\label{sec:artifact}
Our primary correctness label is an LLM judge; a two-rater audit
finds it unbiased in aggregate but at odds with the raters on 26\%
of its hardest cases, and that error propagates to every reported
number (Appendix~\ref{app:labels}). The evaluation is limited to a single
benchmark and database, and EHRSQL is generous to answerability detection
in particular. Its unanswerable questions come from a small set of
short templates, and a question-only baseline already reaches
$0.955$ AUROC, so part of the answerability head's near-ceiling
performance is benchmark structure rather than skill (audit in
Appendix~\ref{app:artifact}); the design absorbs this, since
$\theta$ is set by a wrongly-refused tolerance rather than tuned for
detection accuracy, but detection will be harder in deployment. This mostly affects the gate, which moves $\rs{\lambda}$ by at most $+0.14$ at our 3\% unanswerable rate and carries more of the gain at the benchmark's own 20--33\% (Appendix~\ref{app:abstain}); correctness given answerability is measured on answerable questions only and draws its signal from the trajectory.
We ran no competing confidence method and report one training seed. The confidence heads are also trained per agent on that agent's own trajectories, so a new agent requires a new collection run, and a monitor frozen
at deployment faces an agent whose behavior may drift from the training distribution.
\section{Conclusion and Future Work}
\label{sec:conclusion}

Agentic systems seemingly fail, and in high-stakes settings, the cost of a single delivered mistake can potentially do more damage than one right answer does good. We showed that \sysname{}, two small text classifiers reading the agent's transcript behind utility-derived thresholds, raises the reliability of five text-to-SQL agents at low stakes and, as stakes rise, recognizes on its own when their answers are not reliable enough to deliver. The agent is never modified. Reliability does not have to be built into the agent; our results show that it can be added on top. 

The confidence signal currently only abstains, and the natural next step is to let it act. At training time, the monitor could serve as a cheap verifier that labels preference pairs or supplies rewards for DPO- or GRPO-style fine-tuning of the agent \citep{rafailov2023dpo, shao2024deepseekmath}, though an agent trained against a fixed monitor may learn to look confident rather than to be correct, and this needs to be tested. At inference time, a budget policy could resample a low-confidence trajectory and rerank the candidates instead of stopping it \citep{lightman2023verify, zou2025tattoo}. The heads should also be tested on a new database such as eICU, since a monitor that transfers across schemas is far more useful than one that needs to be rebuilt per database.

\begin{ack}
This research used the DeltaAI advanced computing and data resource,
which is supported by the National Science Foundation (award OAC
2320345) and the State of Illinois. DeltaAI is a joint effort of the
University of Illinois Urbana-Champaign and its National Center for
Supercomputing Applications.
The authors acknowledge the National Artificial Intelligence Research
Resource (NAIRR) Pilot (award NAIRR250187) and NCSA DeltaAI for
contributing to this research result.
\end{ack}

\bibliographystyle{plainnat}
\bibliography{refs}

\appendix
\section{Leakage and artifact battery}
\label{app:artifact}
Recomputed from the collected batch alone
(\texttt{audit\_metrics.csv}): training-free telemetry baselines for
answerability --- mean entropy 0.947; max entropy, mean
log-probability, and the ending-only heuristic (abstained
$\Rightarrow$ unanswerable) in \texttt{audit\_metrics.csv};
question-only TF--IDF 0.955;
full-transcript TF--IDF 0.966; three shuffled-label controls 0.47--0.49;
zero duplicate transcripts across CV folds; BERT-training-loop shuffle
control 0.490 (loss at $\ln 2$). Question-only TF--IDF for
\emph{correctness} reaches 0.689 under the semantic label --- the
question-difficulty prior the trajectory improves upon. In the
training batch, EHRSQL's unanswerable questions are drawn from
${\sim}16$ short topic-label templates (median 17 characters) vs.\
${\sim}100$ full-question answerable templates, explaining the
format signal.

\section{Label decision record}
\label{app:labels}
Strict label: official execution match. Semantic label (primary): GPT
judge (\texttt{GPT-4o-mini}, temperature 0) compares the agent answer's
executed result with the gold query's executed result; the judge prompt
(verbatim, Appendix~\ref{app:prompts}) is instructed to reply with a
single digit. Policy: strict-correct $\Rightarrow$ auto-correct;
abstentions inherit strict; only answered strict-wrong cases are judged.
Outcome: 1{,}294 of 5{,}119 answered questions (25.3\%) flip
strict-wrong $\rightarrow$ semantically-correct (the policy never
shows the judge a strict-correct answer, so reverse flips cannot
occur). Agent correctness is 59.9\% semantic vs.\ 34.6\% strict.
The correctness head trained under each label: $0.889$ under
strict (single earlier run, different hardware configuration; see
\texttt{bert\_prefix\_metrics\_modernbert\_strict.csv}) vs.\
$0.748$ under semantic --- the judge's label is a harder
prediction target, not merely a more lenient one.

\paragraph{Human audit.} Two authors independently re-rated a random
sample of 100 judged verdicts on \texttt{GPT-4o-mini}'s trajectories
(answered, strict-wrong cases, the only ones the judge sees), blind to
the judge and to each other, using the judge's own rule. The raters
agreed on 93/100 (Cohen's $\kappa = 0.86$). On those 93, the judge
matched the human verdict on 69 (74\%, 95\% CI 64--82\%), with errors
in both directions (11 false positives, 13 false negatives), so it
marks slightly fewer answers correct than the raters (38 vs.\ 40) and
does not inflate correctness, including on its own agent's answers.

\section{Encoder decision record}
\label{app:encoder}
Adoption rule: a stronger encoder is adopted if it improves AUROC by
$\ge 0.02$. DistilBERT (512-token window): correctness 0.841,
answerability 0.944, with late-step collapse (0.641 at $t{=}5$).
ModernBERT-base (2{,}048): 0.889 / 0.986, flat per-step profiles ---
the late-step decay was a truncation artifact of the 512-token window,
not a property of deep trajectories. Rule fired on both heads.

\section{Confidence bins}
\label{app:bins}
Delivered-answer accuracy by confidence bin (\texttt{GPT-4o-mini}
shown; every agent's \texttt{confidence\_bins\_*.csv} ships with the
pipeline). On the calibration pool the raw
correctness scores are bimodal and overconfident: 44\% of delivered
answers score $[0.9,1.0]$ raw yet are correct only 71\% of the time,
while 30\% score $[0,0.1]$ and are correct 23\% of the time. After
isotonic calibration, bin accuracy tracks stated confidence, and the
map transfers to held-out dev\_test: $[0.5,0.6] \rightarrow 51\%$
correct, $[0.7,0.8] \rightarrow 77.7\%$, $[0.9,1.0] \rightarrow
100\%$. Only ${\sim}4$\% of dev\_test answers score $\ge 0.8$ under
this head-alone calibration, and the deployed hybrid estimator
delivers 17\% of questions at $\lambda{=}4$ (Table~\ref{tab:oper}):
the agent produces few answers that are genuinely that reliable, and
the system correctly declines the rest.

\section{Operating characteristics}
\label{app:oper}

\begin{table}[h]
  \caption{Operating characteristics (\texttt{GPT-4o-mini},
  dev\_test): the fraction of questions each system answers (cov.)
  and the accuracy of the answers it delivers (acc.), at each stakes
  level. The \texttt{GPT-4o} equivalents appear in the main text
  (Section~\ref{sec:results}).}
  \label{tab:oper}
  \centering
  \small
  \begin{tabular}{l cc cc cc}
    \toprule
    & \multicolumn{2}{c}{$\lambda{=}1$} & \multicolumn{2}{c}{$\lambda{=}4$} & \multicolumn{2}{c}{$\lambda{=}9$} \\
    \cmidrule(lr){2-3} \cmidrule(lr){4-5} \cmidrule(lr){6-7}
    System & Cov. & Acc. & Cov. & Acc. & Cov. & Acc. \\
    \midrule
    vanilla agent    & $0.84$ & $0.48$ & $0.84$ & $0.48$ & $0.84$ & $0.48$ \\
    + Chow delivery  & $0.38$ & $0.73$ & $0.17$ & $0.86$ & $0.03$ & $0.95$ \\
    + early stopping & $0.37$ & $0.73$ & $0.09$ & $0.86$ & $0.01$ & $0.89$ \\
    \bottomrule
  \end{tabular}
\end{table}

\section{Always-abstain comparison and robustness}
\label{app:abstain}
A policy that refuses every question scores the unanswerable fraction
of the pool, $+0.029$ here, at every $\lambda$, with no monitor and no
agent calls. Table~\ref{tab:abstain} reports the full system's gain
over that policy with 95\% intervals from a cluster bootstrap over the
held-out pool's 41 question templates (2{,}000 resamples). The gain is
significant at $\lambda{=}1$ for four agents and indistinguishable
from zero elsewhere, including \texttt{GPT-4o}'s small deficit at
$\lambda{=}4$.

\begin{table}[h]
  \caption{Full-system $\rs{\lambda}$ minus the always-abstain score
  (dev\_test), with 95\% template-cluster bootstrap intervals.}
  \label{tab:abstain}
  \centering
  \footnotesize
  \begin{tabular}{l ccc}
    \toprule
    Agent & $\lambda{=}1$ & $\lambda{=}4$ & $\lambda{=}9$ \\
    \midrule
    \texttt{GPT-4o-mini} & $+0.177\ [+0.088, +0.270]$ & $+0.048\ [-0.017, +0.130]$ & $+0.015\ [-0.014, +0.040]$ \\
    \texttt{GPT-4o}      & $+0.209\ [+0.113, +0.306]$ & $-0.010\ [-0.058, +0.034]$ & $+0.001\ [+0.000, +0.004]$ \\
    \texttt{Llama-3.1-8B} & $+0.029\ [-0.016, +0.081]$ & $+0.000\ [+0.000, +0.002]$ & $+0.000\ [+0.000, +0.001]$ \\
    \texttt{Qwen2.5-Coder-7B} & $+0.075\ [+0.022, +0.135]$ & $+0.001\ [+0.000, +0.003]$ & $+0.001\ [+0.000, +0.003]$ \\
    \texttt{Phi-4-mini}  & $+0.010\ [+0.002, +0.021]$ & $+0.000\ [+0.000, +0.000]$ & $+0.000\ [+0.000, +0.000]$ \\
    \bottomrule
  \end{tabular}
\end{table}

\paragraph{Unanswerable proportion.} The pools merge EHRSQL's train
split, which has no unanswerable questions, with its validation
split, so 3\% of questions are unanswerable against 20--33\% in the
benchmark's own evaluation splits. Reweighting the held-out pool to
20\% / 33\% leaves the $\lambda{=}1$ gain over always abstaining at
$+0.16$ / $+0.12$ for \texttt{GPT-4o} and $+0.15$ / $+0.12$ for
\texttt{GPT-4o-mini}, and raises the gate's share of the
$\lambda{=}1$ gain over the vanilla agent from 4\% to about half for
\texttt{GPT-4o} at 33\%,
since the gate's effect scales with the unanswerable fraction.

\paragraph{Exact threshold.} Replacing $\tau_\lambda$ with the exact
bar $(\lambda + \varepsilon)/(1 + \lambda)$, where $\varepsilon$ is
the calibrated answerability estimate that the question is
unanswerable, changes no full-system entry of Table~\ref{tab:models}
by more than $0.01$.

\paragraph{Strict scoring.} Scoring the same decisions under EHRSQL's
strict execution-match label, the $\lambda{=}1$ gain over always
abstaining falls to $+0.03$ (\texttt{GPT-4o}) and $+0.05$
(\texttt{GPT-4o-mini}), and at $\lambda{=}4$ both fall below it
($-0.11$, $-0.07$). \texttt{GPT-4o}'s delivered answers are 69\% and
79\% correct under the judge at $\lambda{=}1$ and $4$ but 52\% and
66\% under strict matching, and the thresholds were calibrated to the
semantic label, so the size of the gains depends on the judge.

\paragraph{Question-only baselines.} The head scored on the question
alone ($t{=}0$) is each agent's question-only baseline
(Table~\ref{tab:qonly}). The trajectory adds $+0.04$ to $+0.18$ AUROC
to correctness on every agent. For answerability the question alone
carries nearly all of the signal, and on \texttt{Llama-3.1-8B} it is
higher than the delivery-time reading.

\begin{table}[h]
  \caption{Held-out AUROC of each head on the question alone
  ($t{=}0$) and at delivery.}
  \label{tab:qonly}
  \centering
  \small
  \begin{tabular}{l cc cc}
    \toprule
    & \multicolumn{2}{c}{Correctness} & \multicolumn{2}{c}{Answerability} \\
    \cmidrule(lr){2-3} \cmidrule(lr){4-5}
    Agent & $t{=}0$ & Delivery & $t{=}0$ & Delivery \\
    \midrule
    \texttt{GPT-4o-mini} & $0.699$ & $0.787$ & $0.910$ & $0.923$ \\
    \texttt{GPT-4o}      & $0.705$ & $0.762$ & $0.988$ & $0.981$ \\
    \texttt{Llama-3.1-8B} & $0.663$ & $0.707$ & $0.962$ & $0.903$ \\
    \texttt{Qwen2.5-Coder-7B} & $0.742$ & $0.920$ & $0.961$ & $0.959$ \\
    \texttt{Phi-4-mini}  & $0.782$ & $0.921$ & $0.968$ & $0.987$ \\
    \bottomrule
  \end{tabular}
\end{table}

\section{Token-cost efficiency}
\label{app:tokens}
\texttt{GPT-4o-mini} collection measured 2.50M completion tokens
(142/step) and an estimated ${\sim}17$M prompt tokens across
17{,}571 steps. Weighted by generated tokens (exact, from the logged
per-token telemetry), the \texttt{GPT-4o} savings of
Table~\ref{tab:resource} are: gate alone 1\%; early stopping
16\%/47\%/47\% at $\lambda = 1/4/9$ --- slightly above the step
fractions, since eliminated steps skew late and long.

\section{Prompts}
\label{app:prompts}
The agent's system prompt, verbatim; \verb|{schema}| is replaced by the
serialized table schema at run time:

{\footnotesize
\begin{verbatim}
You are a careful clinical data analyst answering questions against a
  hospital SQL database (SQLite).

{schema}

IMPORTANT - database time: this database's "current time" is 2105-12-31
  23:59:00 (records span roughly 2100-2105). You may use datetime('now')
  and relative times; they are automatically mapped to that clock. "This
  year", "current", etc. refer to that timeline, not the real world.

Protocol - reply with exactly one action per turn:
  THOUGHT: <brief reasoning>
  SQL: <one exploratory SQLite query>     (to inspect data; keep LIMIT small)
OR, when you know how to answer:
  THOUGHT: <brief reasoning>
  FINAL_SQL: <the single SQLite query whose result answers the question>
OR, ONLY if the database truly cannot answer this question (the needed
  information is not stored in any table):
  THOUGHT: <brief reasoning>
  FINAL: null

Rules:
- Your goal is ALWAYS to end with FINAL_SQL. Even if you can already see
  the answer in an exploration result, you must still provide the
  FINAL_SQL query that produces it - the query IS your answer.
- An empty result does not mean the question is unanswerable; it may mean
  your query or time window is wrong. Reconsider before giving up.
- Use FINAL: null only when the schema genuinely lacks the information
  (e.g., data the hospital does not record).
Optionally add on its own line:  BEST_SO_FAR: <your current best final
  query, or null>
Notes: use lowercase table/column names as given; do not modify data.
\end{verbatim}
}

The judge prompt, verbatim, sent to \texttt{GPT-4o-mini} at temperature
0 with the system message ``Respond with only the single digit 1 or
0.''; the three placeholders are filled per question, the latter two
with executed results:

{\footnotesize
\begin{verbatim}
You are evaluating an answer against a ground truth reference.

Question: {question}
Ground Truth Answer: {ground_truth}
Predicted Answer: {answer}


RULES
1. If the answer is correct, score it as 1. If the answer is incorrect,
   score it as 0
2. You are not looking for exact string matches, but the answer should
   be very clearly correct
\end{verbatim}
}

\section{Feature definitions}
\label{app:features}
The ten trajectory features (Section~\ref{sec:setup}) are of three
kinds. Token-level telemetry never appears in the transcript, and the
schema checks consult a schema file the head never sees, so neither
can be recovered from the rendered text. The execution aggregates
(empty-result rate, error rate, step count) are visible in the text;
they are included as exact counts that the head would otherwise have
to infer. Purely linguistic signals such as hedging or
self-correction phrasing are excluded: the head already reads those. Every feature is
computable from the first $t$ steps alone, with neutral defaults for
not-yet-observed quantities; the temporal analyses and gate replays
consume exactly this prefix form.

\paragraph{Uncertainty \citep{kadavath2022know, kim2024probgate,
somov2025confidence}.}
\begin{itemize}
  \item \texttt{lp\_mean}: mean over steps of the per-step mean
        chosen-token log-probability.
  \item \texttt{lp\_min}: minimum chosen-token log-probability across
        the whole trajectory.
  \item \texttt{ent\_mean}: mean over steps of the per-step mean token
        entropy, computed over the top-5 alternatives at each position
        (a lower bound on the full entropy).
  \item \texttt{ent\_max}: maximum per-token entropy over the
        trajectory, same truncation; the official \textsc{EHRSQL}
        baseline thresholds this maximum over the full vocabulary
        \citep{lee2022ehrsql}.
\end{itemize}

\paragraph{Execution \citep{wang2018execution, gong2025sqlens}.}
\begin{itemize}
  \item \texttt{empty\_rate}: fraction of executed steps whose query
        returned zero rows.
  \item \texttt{error\_rate}: fraction of executed steps whose query
        raised a SQL error.
  \item \texttt{n\_steps}: number of steps taken so far; doubles as
        the observability index $t$ in the prefix form.
\end{itemize}

\paragraph{Schema grounding \citep{gong2025sqlens}.}
\begin{itemize}
  \item \texttt{failed\_refs}: count, over all logged queries, of
        table or column references that do not resolve against the
        database schema, via an alias-aware parser; identifiers that
        cannot be confidently resolved are conservatively not counted
        as failures.
  \item \texttt{bad\_joins}: count of equality join predicates whose
        column pair is not backed by a declared foreign-key
        relationship.
  \item \texttt{ungrounded\_frac}: fraction of literal values in the
        committed SQL traceable to neither the question text nor any
        previously observed execution result (the database's shifted
        time constants are exempt); ungrounded literals are the
        fabrication signature.
\end{itemize}

\section{Feature analysis: the label-dependent screen}
\label{app:featanalysis}
Each feature is screened for incremental $\Delta$AUROC over BERT-alone
(95\% CI, template-level cluster bootstrap, 1{,}000 resamples over
fixed out-of-fold predictions). Under the semantic label, the
gradient-boosted hybrid improves on BERT-alone by $+0.028$ AUROC
(CI $[+0.011, +0.047]$); the failed-reference count survives the
per-feature screen, while inside the hybrid the execution family
carries the weight: removing it costs $-0.015$, and the uncertainty
and schema families add little.
Under the strict label the same screen yields no survivors: a strong
text reader saturates a clean execution-match target, while
externally observed execution events add signal when the target is
the noisier semantic judgment. On the calibration pool the hybrid
scores $0.780$ vs.\ $0.746$ for BERT-alone, so the pre-registered
selection rule deploys the hybrid at delivery. The selection
replicates on all five agents (Table~\ref{tab:models}), with the
execution family load-bearing every time (knockouts $-0.009$ to
$-0.038$); the margin is largest on the agent whose transcript is
least legible to the text reader (\texttt{Llama-3.1-8B}, $0.748$ vs
$0.691$), the pattern expected if features compensate where text
reading fails. The per-feature survivor matrix is
Table~\ref{tab:featscreen}; per-agent screen details, including
each feature's $\Delta$AUROC and CI, ship in
\texttt{hybrid\_feature\_screen.csv}.

\begin{table}[h]
  \caption{\textbf{Which features individually survive the screen,
  per agent} (train pool, out-of-fold; \checkmark{} = incremental
  $\Delta$AUROC over BERT-alone with 95\% cluster-bootstrap CI
  excluding zero; not a held-out result). Survivor counts are
  highest for the two agents the head reads worst, though five agents
  cannot establish a trend; only aggregate rulings are agent-general.}
  \label{tab:featscreen}
  \centering
  \footnotesize
  \setlength{\tabcolsep}{3pt}
  \resizebox{\linewidth}{!}{%
  \begin{tabular}{l cccccccccc c}
    \toprule
    Agent & \rotatebox{55}{\texttt{lp\_mean}} & \rotatebox{55}{\texttt{lp\_min}} & \rotatebox{55}{\texttt{ent\_mean}} & \rotatebox{55}{\texttt{ent\_max}} & \rotatebox{55}{\texttt{empty\_rate}} & \rotatebox{55}{\texttt{error\_rate}} & \rotatebox{55}{\texttt{n\_steps}} & \rotatebox{55}{\texttt{failed\_refs}} & \rotatebox{55}{\texttt{bad\_joins}} & \rotatebox{55}{\texttt{ungrounded\_frac}} & \rotatebox{55}{Survivors} \\
    \midrule
    \texttt{GPT-4o-mini} & $\times$ & $\times$ & $\times$ & $\times$ & $\times$ & $\times$ & $\times$ & \checkmark & $\times$ & $\times$ & 1 \\
    \texttt{GPT-4o}      & \checkmark & \checkmark & \checkmark & \checkmark & \checkmark & $\times$ & \checkmark & $\times$ & \checkmark & $\times$ & 7 \\
    \texttt{Llama-3.1-8B} & $\times$ & $\times$ & $\times$ & $\times$ & \checkmark & \checkmark & \checkmark & \checkmark & \checkmark & $\times$ & 5 \\
    \texttt{Qwen2.5-Coder-7B} & $\times$ & $\times$ & $\times$ & $\times$ & $\times$ & $\times$ & $\times$ & $\times$ & $\times$ & $\times$ & 0 \\
    \texttt{Phi-4-mini}  & $\times$ & $\times$ & $\times$ & $\times$ & $\times$ & \checkmark & $\times$ & \checkmark & $\times$ & $\times$ & 2 \\
    \bottomrule
  \end{tabular}}
\end{table}

\section{Per-agent exhibits}
\label{app:peragent}
Every single-agent exhibit in the main text is replicated for all
five agents (Table~\ref{tab:models}); the released pipeline ships
each agent's complete results CSVs (system ladders, risk--coverage
sweeps, confidence bins, per-step metrics). Figure~\ref{fig:results}'s \texttt{GPT-4o} panels are reproduced
below for each remaining agent, alongside the
\texttt{GPT-4o-mini} exhibits (the development agent on which the
labels, audits, and feature analyses were built).

\begin{table}[h]
  \caption{System ladder for \texttt{GPT-4o-mini} (dev\_test,
  2{,}100 questions, evaluated once): each decision component's
  reliability and compute contribution.}
  \label{tab:rs-mini}
  \centering
  \small
  \begin{tabular}{l ccc c}
    \toprule
    System & $\rs{1}$ & $\rs{4}$ & $\rs{9}$ & Steps Saved ($\lambda{=}1/4/9$) \\
    \midrule
    vanilla agent      & $-0.02$ & $-1.34$ & $-3.53$ & --- \\
    + exit gate        & $-0.01$ & $-1.29$ & $-3.43$ & $2\%$ \\
    + Chow delivery    & $+0.21$ & $+0.08$ & $+0.04$ & $2\%$ \\
    + early stopping   & $+0.20$ & $+0.06$ & $+0.03$ & $28\%\,/\,64\%\,/\,70\%$ \\
    \bottomrule
  \end{tabular}
\end{table}

\begin{figure}[H]
  \centering
  \includegraphics[width=\linewidth]{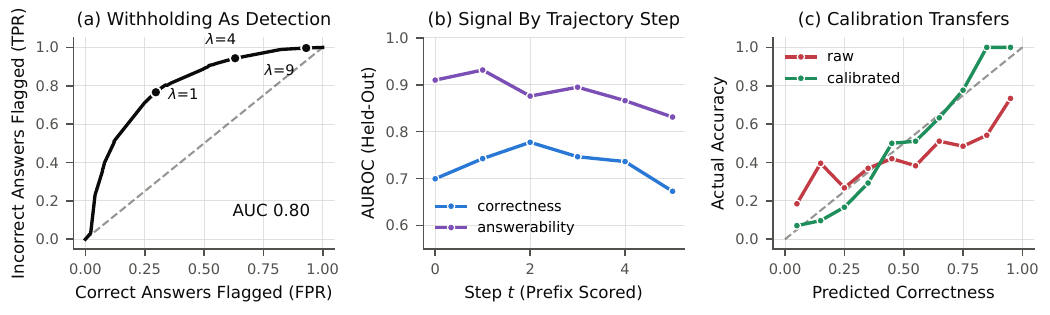}
  \caption{The Figure~\ref{fig:results} panels for \texttt{GPT-4o-mini}
  (dev\_test; withhold-classifier AUC $0.80$).}
  \label{fig:results-mini}
\end{figure}

\begin{figure}[H]
  \centering
  \includegraphics[width=\linewidth]{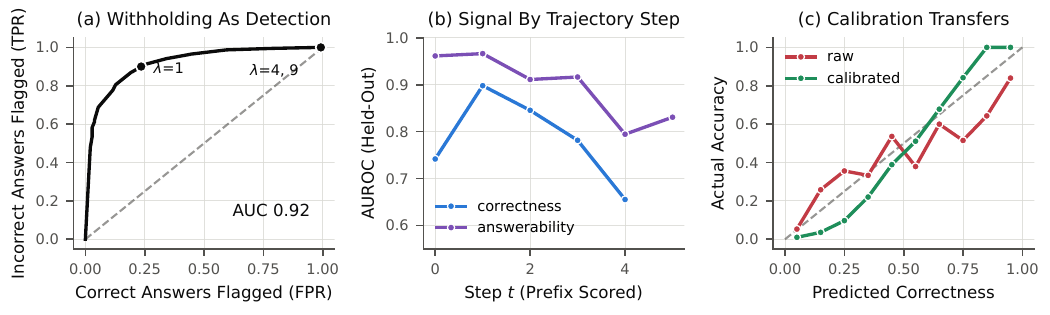}
  \caption{The Figure~\ref{fig:results} panels for \texttt{Qwen2.5-Coder-7B}
  (dev\_test; withhold-classifier AUC $0.92$).}
  \label{fig:results-qwen}
\end{figure}

\begin{figure}[H]
  \centering
  \includegraphics[width=\linewidth]{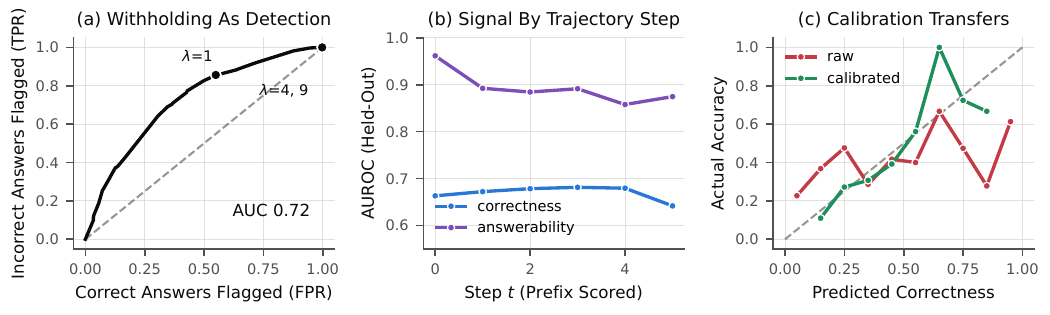}
  \caption{The Figure~\ref{fig:results} panels for \texttt{Llama-3.1-8B}
  (dev\_test; withhold-classifier AUC $0.72$).}
  \label{fig:results-llama}
\end{figure}

\begin{figure}[H]
  \centering
  \includegraphics[width=\linewidth]{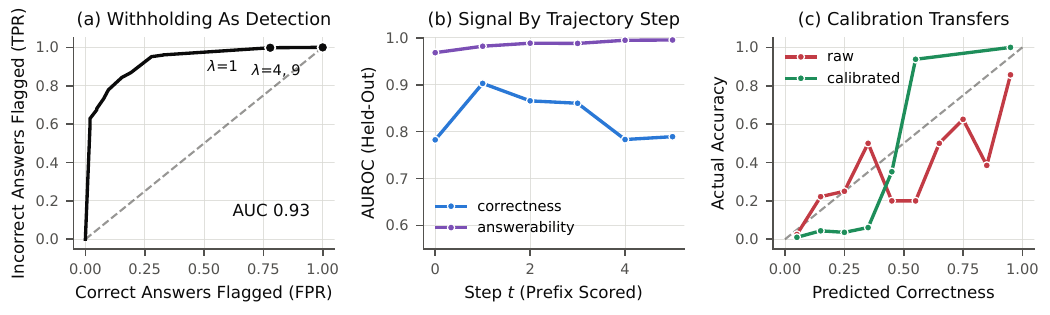}
  \caption{The Figure~\ref{fig:results} panels for \texttt{Phi-4-mini}
  (dev\_test; withhold-classifier AUC $0.93$).}
  \label{fig:results-phi}
\end{figure}

\begin{table}[h]
  \caption{Early-stop mode per agent (dev\_test): $\rs{\lambda}$ with
  sequential stopping, and the fraction of agent steps never
  executed.}
  \label{tab:es-agents}
  \centering
  \small
  \begin{tabular}{l ccc ccc}
    \toprule
    & \multicolumn{3}{c}{Early-Stop $\rs{\lambda}$} & \multicolumn{3}{c}{Steps Saved} \\
    \cmidrule(lr){2-4} \cmidrule(lr){5-7}
    Agent & $1$ & $4$ & $9$ & $1$ & $4$ & $9$ \\
    \midrule
    \texttt{GPT-4o-mini} & $+0.20$ & $+0.06$ & $+0.03$ & $28\%$ & $64\%$ & $70\%$ \\
    \texttt{GPT-4o}      & $+0.24$ & $+0.02$ & $+0.03$ & $13\%$ & $42\%$ & $42\%$ \\
    \texttt{Llama-3.1-8B} & $+0.06$ & $+0.03$ & $+0.03$ & $14\%$ & $53\%$ & $53\%$ \\
    \texttt{Qwen2.5-Coder-7B} & $+0.10$ & $+0.03$ & $+0.03$ & $22\%$ & $62\%$ & $62\%$ \\
    \texttt{Phi-4-mini}  & $+0.04$ & $+0.03$ & $+0.03$ & $15\%$ & $32\%$ & $32\%$ \\
    \bottomrule
  \end{tabular}
\end{table}

\begin{table}[h]
  \caption{Step economics of unanswerable questions per agent (train
  pool, mean steps): fabricating an answer vs.\ correctly refusing
  (an explicit abstention; runs that hit the 8-step cap are
  excluded) vs.\ a typical answerable question. Four of five agents work harder
  on the fabrication than on the correct refusal;
  \texttt{Phi-4-mini} is the exception, fabricating as fast as it
  refuses.}
  \label{tab:fab-steps}
  \centering
  \small
  \begin{tabular}{l ccc}
    \toprule
    Agent & Fabrication & Correct Refusal & Typical Answerable \\
    \midrule
    \texttt{GPT-4o-mini} & $3.29$ & $2.12$ & $2.86$ \\
    \texttt{GPT-4o}      & $3.35$ & $2.12$ & $3.01$ \\
    \texttt{Llama-3.1-8B} & $5.26$ & $4.46$ & $4.97$ \\
    \texttt{Qwen2.5-Coder-7B} & $2.25$ & $1.68$ & $1.96$ \\
    \texttt{Phi-4-mini}  & $1.75$ & $1.70$ & $3.36$ \\
    \bottomrule
  \end{tabular}
\end{table}

The \texttt{Phi-4-mini} reversal is informative. For the other four
agents fabrication is \emph{effortful}: the agent explores, retrieves
something adjacent-looking, and rationalizes it into an answer, so
the fabrication carries the cost of the investigation that produced
its misleading evidence (Figure~\ref{fig:motivation}, Q1).
\texttt{Phi-4-mini} shows no such signature: both of its outcomes on
unanswerable questions are near-instant ($1.75$ vs.\ $1.70$ steps)
while its rare attempts to explore them run to the step cap,
consistent with fabrication by reflex --- committing from the
question's surface form without gathering evidence --- rather than by
rationalization. The same shallowness is what makes its transcripts
the easiest to monitor (Table~\ref{tab:models}): a reflex fabrication
leaves no competent-looking trail for the correctness head to be
fooled by.

\paragraph{External reference point.}

For external context on the shared scale of
\citet{broecker2026idk}: this agent's $\lambda{=}1$ lift
($-0.02 \rightarrow +0.21$) is comparable to the ${\approx}{+}0.2$
IDK-score gain their filter reports for a \texttt{Llama-3.2-3B}
agent on APIGen ($0.34 \rightarrow 0.54$). The IDK score counts
abstentions as $0$, so it coincides with $\rs{1}$ only on answerable
questions. Early
stopping prices its one real trade on this agent: at $\lambda{=}4$
it buys 64\% of steps for $0.02$ $\rs{4}$, from a stop threshold
tuned where calibration-pool coverage was thin; the timed serial
rerun measures $2.2$ seconds per step for this agent.

\section{Example rendered transcript}
\label{app:transcript}
One full rendering, exactly as the heads read it (question
\texttt{d7c5f51c}, train pool): an unanswerable question (department
scoping is not represented in the schema) on which the agent performs
seven steps of plausible-looking exploration before abstaining on its
own. The answerability head is not distracted by the productive-looking
activity: its out-of-fold $\pans$ stays below $10^{-3}$ at every step
including $t{=}0$, so the exit gate would have refused before step 1,
saving all seven steps.

{\footnotesize
\begin{verbatim}
QUESTION: Can you provide me with all the prescriptions received by
  patient 16879 in another department?
STEP 1 THOUGHT: I need to find the prescriptions for patient 16879 and
  check if there are any records indicating that these prescriptions
  were given in a department other than the one they were admitted to.
  However, the prescriptions table does not directly include department
  information. I will need to join it with the admissions table to
  determine the department.
STEP 1 SQL: SELECT p.* FROM prescriptions p JOIN admissions a ON
  p.subject_id = a.subject_id AND p.hadm_id = a.hadm_id WHERE
  p.subject_id = 16879 AND a.admission_location != 'ICU'
STEP 1 RESULT: empty (0 rows)
STEP 2 THOUGHT: The result indicates that there are no prescriptions
  found for patient 16879 in the current query setup. It's possible
  that the admission location I used ('ICU') was incorrect or too
  restrictive. I will try checking all prescriptions for patient 16879
  without restricting the admission location and then look for any
  department-related information in the admissions table.
STEP 2 SQL: SELECT p.* FROM prescriptions p JOIN admissions a ON
  p.subject_id = a.subject_id AND p.hadm_id = a.hadm_id WHERE
  p.subject_id = 16879
STEP 2 RESULT: empty (0 rows)
STEP 3 THOUGHT: Since there are no prescriptions found for patient
  16879, I need to verify if this patient has any admissions recorded.
  If there are no admissions, it would explain the absence of
  prescriptions. I will query the admissions table to check for any
  records related to patient 16879.
STEP 3 SQL: SELECT * FROM admissions WHERE subject_id = 16879
STEP 3 RESULT: 1 rows x 13 cols; sample [['313', '16879', '123812',
  '2103-09-13 10:21:00', '2103-10-07 18:00:00', 'elective', 'phys
  referral/normal deli', 'home health care', 'medicaid', 'None',
  'married', 'asian', '31']]
STEP 4 THOUGHT: Patient 16879 has one admission recorded, which
  indicates that they were admitted on 2103-09-13 and discharged on
  2103-10-07. Since there are no prescriptions linked to this patient
  in the previous queries, I will check if there are any procedures or
  diagnoses recorded during this admission that may provide additional
  context or confirm the patient's treatment during their hospital
  stay.
STEP 4 SQL: SELECT * FROM procedures_icd WHERE subject_id = 16879
STEP 4 RESULT: 7 rows x 5 cols; sample [['627', '16879', '123812',
  '9904', '2103-10-07 18:00:00'], ['628', '16879', '123812', '3891',
  '2103-10-07 18:00:00'], ['629', '16879', '123812', '9955',
  '2103-10-07 18:00:00']]
STEP 5 THOUGHT: Patient 16879 has several procedures recorded during
  their admission. However, I still need to find any prescriptions that
  may have been associated with these procedures or the admission
  itself. I will check if there are any diagnoses associated with
  patient 16879 during the same admission.
STEP 5 SQL: SELECT * FROM diagnoses_icd WHERE subject_id = 16879
STEP 5 RESULT: 7 rows x 5 cols; sample [['3053', '16879', '123812',
  '42833', '2103-09-13 10:21:00'], ['3054', '16879', '123812', '4280',
  '2103-09-13 10:21:00'], ['3055', '16879', '123812', '4019',
  '2103-09-13 10:21:00']]
STEP 6 THOUGHT: Patient 16879 has several diagnoses recorded during
  their admission. While I have gathered information about their
  admissions, procedures, and diagnoses, I still have not found any
  prescriptions associated with this patient. Since I have confirmed
  the patient has an admission and procedures, I will now check if
  there are any prescriptions linked to their admission by looking for
  prescriptions associated with their HADM_ID.
STEP 6 SQL: SELECT * FROM prescriptions WHERE hadm_id = 123812
STEP 6 RESULT: empty (0 rows)
STEP 7 THOUGHT: The absence of prescriptions for patient 16879, despite
  having an admission and procedures, suggests that this patient may
  not have received any medications during their hospital stay. This
  could be due to the nature of their treatment or other clinical
  decisions made at that time. Therefore, I will conclude that there
  are no prescriptions received by patient 16879 in any department.
FINAL: null (abstained)
\end{verbatim}
}

\section{Reproducibility and cost}
\label{app:repro}
Seed 42 is the run of record for every agent (a one-time three-seed
check on \texttt{GPT-4o-mini}, retained in the released
\texttt{seed\_variance.csv}, put the terminal-AUROC SD at
${\pm}0.009$; seed reruns were dropped thereafter as not worth the
compute). \texttt{transformers==4.56.2}
pinned; the decision layer runs in one environment with
\texttt{scikit-learn==1.6.1}, since the gradient-boosted hybrid's fit
varies across versions; collection gaps leave 2{,}099 held-out
questions for \texttt{GPT-4o} and \texttt{Qwen2.5-Coder-7B} and
2{,}100 for \texttt{GPT-4o-mini}, and each agent's metrics are
computed over the questions it completed; template-to-question pool manifest shipped with the code;
two-tier reproducibility (exact artifacts released; retraining
reproduces statistically). Compute and cost are per-agent: the primary
\texttt{GPT-4o-mini} run took approximately 2 GPU-hours per
head-training run and roughly \$4 of API spend for collection plus
labeling (estimates; not instrumented). Open-source agents spend API
only on the judge (the agent itself is served on a local GPU); a
flagship \texttt{GPT-4o} run is substantially more expensive
(${\sim}\$110$: 30M prompt + 3.6M completion tokens at list prices, computed from the logged telemetry). Confidence intervals via template-level
cluster bootstrap (1{,}000 resamples), not seed reruns.

\end{document}